\documentclass{article}

\usepackage{microtype}
\usepackage{graphicx}
\usepackage{subfigure}
\usepackage{booktabs} 

\usepackage{hyperref}

\usepackage[accepted]{icml2024}

\usepackage{amsmath}
\usepackage{amssymb}
\usepackage{mathtools}
\usepackage{amsthm}

\usepackage[capitalize,noabbrev]{cleveref}

\theoremstyle{plain}

\theoremstyle{definition}

\theoremstyle{remark}

\usepackage[textsize=tiny]{todonotes}

\icmltitlerunning{Submission and Formatting Instructions for ICML 2024}

\begin{document}

\twocolumn[
\icmltitle{Generative Autoencoding of Dropout Patterns}



\icmlsetsymbol{equal}{*}

\begin{icmlauthorlist}
\icmlauthor{Shunta Maeda}{comp}
\end{icmlauthorlist}

\icmlaffiliation{comp}{Uchr Technology, Tokyo, Japan}

\icmlcorrespondingauthor{Shunta Maeda}{shunta@uchrtech.com}

\icmlkeywords{Machine Learning, ICML}

\vskip 0.3in
]



\printAffiliationsAndNotice{\icmlEqualContribution} 

\begin{abstract}
We propose a generative model termed Deciphering Autoencoders.
In this model, we assign a unique random dropout pattern to each data point in the training dataset and then train an autoencoder to reconstruct the corresponding data point using this pattern as information to be encoded.
Even if a completely random dropout pattern is assigned to each data point regardless of their similarities, a sufficiently large encoder can smoothly map them to a low-dimensional latent space to reconstruct individual training data points.
During inference, using a dropout pattern different from those used during training allows the model to function as a generator.
Since the training of Deciphering Autoencoders relies solely on reconstruction error, it offers more stable training compared to other generative models.
Despite their simplicity, Deciphering Autoencoders show sampling quality comparable to DCGAN on the CIFAR-10 dataset. 
Code: \url{https://github.com/shuntama/deciphering-autoencoders}
\end{abstract}

\section{Introduction}
\label{submission}

Recent advancements in generative image models have primarily focused on decomposing the generative process into incremental steps~\citep{ho2020denoising, delbracio2023inversion}.
While highly effective, this approach is not without its challenges.
Iterative models can result in extended computation times and increased sensitivity to hyperparameters, complicating the training process.
Although models like Generative Adversarial Networks (GANs)~\citep{goodfellow2014generative} and Variational Autoencoders offer single-step inferences as potential alternatives, they too encounter issues of training instability.
Such instability can hinder scalability, which is essential for the success of deep learning models.
Against this backdrop, our study explores simple autoencoders~\citep{vincent2008extracting, bengio2013generalized}, with the aim of highlighting and advancing stable, scalable single-step generative models.

Variational Autoencoders (VAEs)~\citep{kingma2013auto} are generative models that can directly sample from a decoder.
By assuming a prior distribution for training in the latent space, VAEs can generate new samples by drawing from this distribution.
VAEs offer more stable training than GANs and enable faster sampling than other generative models, such as autoregressive and diffusion models.
However, the constraints imposed on the latent space can compromise the quality of the generated samples.
Furthermore, balancing the reconstruction error and the KL divergence term during training can be a practical challenge, often requiring specific adjustments to avoid issues like over-regularization and posterior collapse~\citep{van2017neural}.

Regularized Autoencoders (RAEs)~\citep{ghosh2019variational} have been proposed to address these inherent issues in VAEs.
They do not use the KL divergence term but instead introduce a regularization term to prevent overfitting while preserving smoothness in the latent space.
However, due to the lack of control over the learned latent space distribution, RAEs require a posterior density estimation step for sampling.

In this paper, we propose a deterministic generative autoencoding framework named Deciphering Autoencoders that does not require assumptions about the latent space distribution nor posterior density estimation.
Our approach commences by assigning a unique, randomly generated pattern to each data point in the training dataset (ciphering).
This pattern is then encoded using an encoder-decoder network to reconstruct the corresponding data point (deciphering).
The objective function relies solely on the reconstruction error, promoting highly stable training.
For sampling, we generate new random patterns from the distribution used during training.
These patterns are then encoded to produce fresh samples.
We have observed that utilizing dropout patterns as random patterns for encoding enhances the model's training. Additionally, we propose a structural implicit regularization technique to mitigate overfitting.
Deciphering Autoencoders exhibit sampling quality comparable to that of DCGAN~\citep{radford2015unsupervised} on the CIFAR-10 dataset~\citep{krizhevsky2009learning}.

\section{Deciphering Autoencoders}\label{sec:formulation}

Our first step involves generating unique random patterns $\{z_i\}_{i=1}^N \in \mathcal{Z}$ corresponding to each element of our training dataset $\{x_i\}_{i=1}^N \in \mathcal{X}$.
This process results in forming a pair of datasets $\{(x_i, z_i)\}_{i=1}^N$.
Here, $x_i$ is an image with dimensions $3 \times h \times w$.
The random patterns $z_i$ can be any that can be encoded by the encoder network.
The whole of encoder-decoder network as generator $g_\theta: \mathcal{Z} \rightarrow \mathcal{X}$ learns the parameters $\theta$ by minimizing a following reconstruction error

\begin{equation}
  \frac{1}{N} \sum_{i=1}^N d(g_\theta(z_i), x_i).
  \label{eq01}
\end{equation}

In this study, we employed LPIPS (Learned Perceptual Image Patch Similarity) metric~\citep{zhang2018unreasonable} as the distance measure \footnote{While it is feasible to use MSE as the distance function for training, it results in blurry generated images. Therefore, we adopted LPIPS to improve image quality.}.
Sampling is performed by generating new random patterns from the same distribution used during training and inputting these into the model.
We will refer to this model as Deciphering Autoencoders.
This framework can be conceptualized as ciphering each $x_i$ into $z_i$, and subsequently deciphering $z_i$ to reconstruct $x_i$ through the training.

\begin{figure}[t]
\vskip 0.2in
\begin{center}
\centerline{\includegraphics[width=0.88\columnwidth]{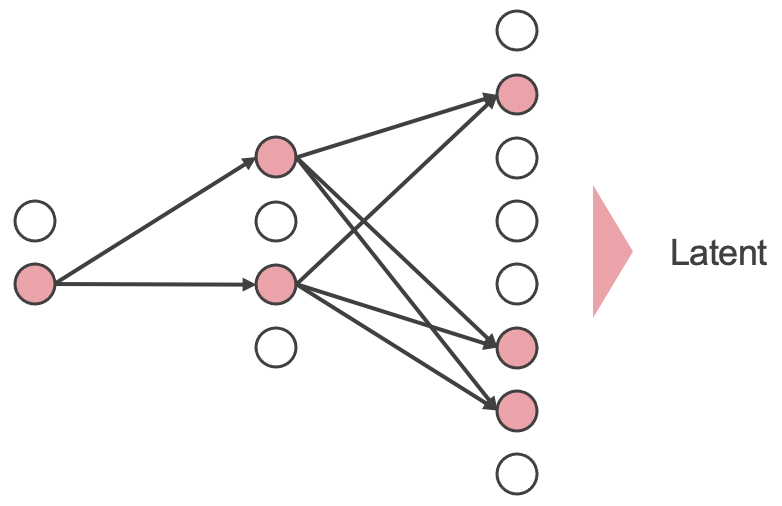}}
\caption{Conceptual diagram of
the encoder in Deciphering Autoencoders.}
\label{fig_encoder}
\end{center}
\vskip -0.2in
\end{figure}

\paragraph{Configuration of $z_i$}
We integrated channel-wise dropout layers after each encoder hierarchy and adopted this dropout patterns unique to each $x_i$ as $z_i$.
Figure~\ref{fig_encoder} presents a conceptual diagram of the encoder in Deciphering Autoencoders.
In the diagram, only the channels indicated in red are activated, and the pattern of these activated channels is assigned as unique to each data point.
By utilizing an encoder with sufficiently large parameters, we anticipate that random activation patterns can be smoothly organized in a low-dimensional latent space.

\paragraph{Regularization}
To enhance the quality of the generated samples, we have introduced a geometric regularization technique.
The proposed geometric regularization involves applying a geometric transformation $\mathcal{T}$ to $x_i$ using random transformation parameters $r$.
Subsequently, we input both $r$ and $z_i$ into the model to decode the transformed $x_i$.
As a result, Equation~\ref{eq01} is updated as follows:

\begin{equation}
  \frac{1}{N} \sum_{i=1}^N d(g_\theta(z_i, r), \mathcal{T}(x_i, r)).
  \label{eq02}
\end{equation}

In this study, we employed horizontal spatial shift as the chosen geometric transformation.
Through this regularization approach, we anticipate the encoder to encode more abstract features of the images that are independent of their spatial positions.
Additionally, apart from the geometric regularization, we also apply regularization through the use of a high learning rate and substantial weight decay during the optimization process.

\paragraph{Model architecture}
We employed an encoder-decoder network that incorporates residual blocks~\citep{he2016deep} with batch normalization~\citep{ioffe2015batch}.
In the decoder, group convolution is utilized as needed to reduce the number of parameters \footnote{To smoothly map random activation patterns to a low-dimensional latent space, the encoder requires a sufficiently large number of parameters. However, to avoid overfitting, the number of parameters in the decoder should be kept to a necessary minimum.}.
The spatial shift information of the geometric regularization is processed by a Multi-Layer Perceptron (MLP) and is then input to the decoder alongside the latent variables.
We emphasize that the number of active layers in the channel-wise dropout used as $z_i$ is not determined stochastically; instead, it is configured so that a fixed number of channels are active in each layer.
In this work, the number of channels in each hierarchy of the encoder is 128, 256, and 512, and the number of active channels that are not suppressed by the channel-wise dropout is set to 1, 4, and 16, respectively.
With this configuration, the possible number of $z_i$ patterns is $\binom{128}{1} \times \binom{256}{4} \times \binom{512}{16} \simeq 1.88 \times 10^{40}$, which is sufficiently large compared to the size of the training dataset.

\section{Results}

\begin{figure*}[t]
\vskip 0.2in
\begin{center}
\centerline{\includegraphics[width=2.07\columnwidth]{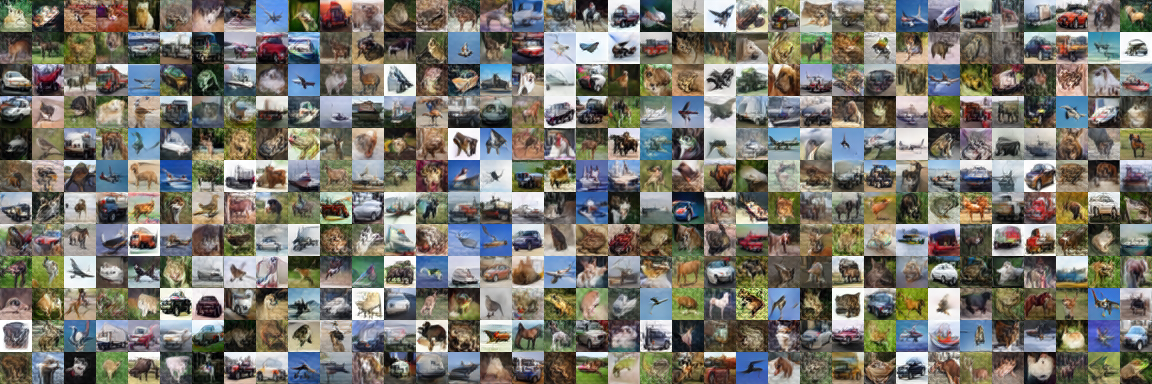}}
\caption{Randomly generated CIFAR-10 results.}
\label{fig_samples}
\end{center}
\vskip -0.2in
\end{figure*}

\paragraph{Implementation}
To train the model, we employed the AdamW optimizer~\citep{loshchilov2017decoupled} with a learning rate of 2e-3 and a batch size of 256 for a total of 1000 epochs.
During this training process, weight decay was linearly warmed up from 0.0 to 0.08 over initial 400 epochs.
Notably, only the MLP module, which takes the shift amount for geometric regularization as input, had a lower learning rate set to 2e-4.
The maximum shift amount for geometric regularization was limited to 8 pixels.
After the initial 1000 epochs of training, geometric regularization was disabled, and training was extended for an additional 2000 epochs.
Model evaluation was performed using model weights with an exponential moving average at a decay rate of 0.99995.
The implementation was carried out using PyTorch~\citep{paszke2019pytorch}, and all experiments were conducted on a single NVIDIA A4000 GPU.
The complete model training process required approximately 30 hours.
It is worth noting that batch normalization layers were inserted after all convolution and transposed convolution layers except for the final layer of the network, and these batch normalization layers were essential for the successful training of the model.

\begin{table}[t]
\caption{Quantitative results for CIFAR-10 train and test dataset.}
\label{main_results}
\vskip 0.15in
\begin{center}
\begin{tabular}{lll}
    \toprule
            & FID ($\downarrow$)    & IS ($\uparrow$)\\
    \midrule
    train   & 39.02                 & 6.84 \\
    test    & 42.73                 & 6.77 \\
    \bottomrule
\end{tabular}
\end{center}
\vskip -0.1in
\end{table}

\paragraph{CIFAR-10}
Table~\ref{main_results} presents the results of unconditional generation using Deciphering Autoencoders trained on CIFAR-10.
The evaluation metrics utilized are the Frechet Inception Distance (FID)~\citep{heusel2017gans} and Inception Score (IS)~\citep{salimans2016improved} \footnote{We calculated FID using pytorch-fid (\url{https://github.com/mseitzer/pytorch-fid}) and IS using torch-fidelity (\url{https://github.com/toshas/torch-fidelity}).}.
The performance achieved by Deciphering Autoencoders is comparable to that of DCGAN.
Figure~\ref{fig_samples} showcases randomly generated images.

In the formulation described in Section~\ref{sec:formulation}, completely random dropout patterns are assigned to each data point.
Interestingly, we observed a slight improvement in performance when the training data were pre-clustered.
Clustering information was conveyed to the model by selectively activating channels in the first dropout layer of the encoder.
We employed the k-means method for clustering. Table~\ref{numclusters} presents the relationship between the number of clusters and model performance \footnote{For this experiment, we conducted training for only 1000 epochs without extending fine-tuning.}.
Performance improves as the number of clusters increases up to 32, but no further improvement is observed beyond that.
Notably, in other experiments within this paper, the number of clusters is consistently set to 32.

\begin{table}[t]
\caption{Quantitative results for CIFAR-10 train and test dataset.}
\label{numclusters}
\vskip 0.15in
\begin{center}
\begin{tabular}{lll}
    \toprule
    number of clusters  & FID ($\downarrow$)    & IS ($\uparrow$)\\
    \midrule
    1 (w/o clustering)  & 48.72                 & 5.96 \\
    8                   & 42.83                 & 6.34 \\
    16                  & 45.88                 & 6.36 \\
    32                  & 44.15                 & 6.63 \\
    64                  & 44.39                 & 6.58 \\
    \bottomrule
\end{tabular}
\end{center}
\vskip -0.1in
\end{table}

Finally, we conducted training without employing geometric regularization to assess its effect.
The results yielded $\rm{FID} = 47.69$ and $\rm{IS} = 6.12$, indicating a decrease in performance compared to when geometric regularization was employed ($\rm{FID} = 42.73$, $\rm{IS} = 6.77$).

\begin{figure}[t]
\vskip 0.2in
\begin{center}
    \includegraphics[width=0.8\columnwidth]{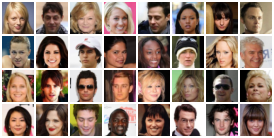}
    \subfigure[Reconstructed samples]{\includegraphics[width=0.8\columnwidth]{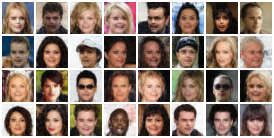}}
    \subfigure[Randomly generated samples]{\includegraphics[width=0.8\columnwidth]{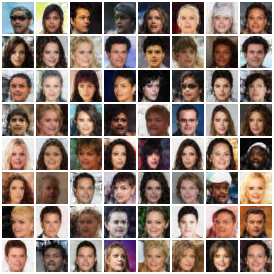}}
    \caption{Qualitative analysis on CelebA dataset.}
    \label{fig_celeba}
\end{center}
\vskip -0.2in
\end{figure}

\paragraph{CelebA}
To verify the generalizability of the proposed method, we conducted experiments using the CelebA dataset~\citep{liu2015deep}.
We utilized 162,770 images from the CelebA training set.
As a preprocessing step, we cropped the central $160\times160$ pixels of each image and resized them to $32\times32$ pixels.
Additionally, we modified some training conditions to suit the dataset: latent dimension was set to 256, and image shift to 0.
Figure~\ref{fig_celeba} shows the results after 650 epochs of training \footnote{We trained the model with a learning rate of 2e-3 for 500 epochs, then reduced the learning rate by a factor of 10 for 100 epochs, and again by another factor of 10 for the final 50 epochs. In the case of CIFAR-10, lowering the learning rate led to overfitting, whereas no such issue was observed with CelebA.}.
Here, we present only qualitative results.
These results demonstrate that our proposed method is effective even for datasets with different domains and data sizes.

\section{Related Works}

Ghosh et al.~\citep{ghosh2019variational} proposed that VAEs could be viewed as Autoencoders with Gaussian noise added to the decoder input.
They suggested that this concept could be substituted with decoder regularization and, consequently, introduced a simpler deterministic framework called Regularized Autoencoders.
However, this model sacrifices the ability to sample from its prior distribution.
To address this limitation, they incorporated an ex-post density estimation step for generating new sample.

Saseendran et al.~\citep{saseendran2021shape} extended the work of Ghosh et al.~\citep{ghosh2019variational} by introducing a deterministic regularization scheme that efficiently shapes the latent space of the model during training.
As a result, the latent distribution is guided toward an expressive predetermined prior, eliminating the need for an ex-post density estimation step.

Bojanowski et al.~\citep{bojanowski2017optimizing} proposed Generative Latent Optimization (GLO), a framework to train generators only with simple reconstruction losses.
In this framework, a set of random vectors is prepared to be paired with a set of training images.
The initialized random vectors are jointly optimized with the generator to be modified into the proper vectors for each image.
Unlike GLO, in Deciphering Autoencoders, a random dropout pattern is used to represent each data point, and these patterns remain constant throughout the training. The encoder takes over the optimization of the latent space.

\section{Conclusion}

Deciphering Autoencoders is a deterministic generative autoencoding framework that provides stable training solely based on a reconstruction error.
Despite its simplicity, it demonstrates image generation performance comparable to the initially proposed GANs.
Notably, the theoretical understanding of why Deciphering Autoencoders functions effectively as a generative model remains unclear, and the exploration of the training protocol and model structure is still insufficient.
Addressing these improvements and achieving theoretical clarifications will be a challenge for future research.


\nocite{langley00}

\bibliography{ref}

\begin{thebibliography}{20}
\providecommand{\natexlab}[1]{#1}
\providecommand{\url}[1]{\texttt{#1}}
\expandafter\ifx\csname urlstyle\endcsname\relax
  \providecommand{\doi}[1]{doi: #1}\else
  \providecommand{\doi}{doi: \begingroup \urlstyle{rm}\Url}\fi

\bibitem[Bengio et~al.(2013)Bengio, Yao, Alain, and Vincent]{bengio2013generalized}
Bengio, Y., Yao, L., Alain, G., and Vincent, P.
\newblock Generalized denoising auto-encoders as generative models.
\newblock \emph{NeurIPS}, 2013.

\bibitem[Bojanowski et~al.(2017)Bojanowski, Joulin, Lopez-Paz, and Szlam]{bojanowski2017optimizing}
Bojanowski, P., Joulin, A., Lopez-Paz, D., and Szlam, A.
\newblock Optimizing the latent space of generative networks.
\newblock \emph{arXiv preprint arXiv:1707.05776}, 2017.

\bibitem[Delbracio \& Milanfar(2023)Delbracio and Milanfar]{delbracio2023inversion}
Delbracio, M. and Milanfar, P.
\newblock Inversion by direct iteration: An alternative to denoising diffusion for image restoration.
\newblock \emph{arXiv preprint arXiv:2303.11435}, 2023.

\bibitem[Ghosh et~al.(2019)Ghosh, Sajjadi, Vergari, Black, and Sch{\"o}lkopf]{ghosh2019variational}
Ghosh, P., Sajjadi, M.~S., Vergari, A., Black, M., and Sch{\"o}lkopf, B.
\newblock From variational to deterministic autoencoders.
\newblock \emph{arXiv preprint arXiv:1903.12436}, 2019.

\bibitem[Goodfellow et~al.(2014)Goodfellow, Pouget-Abadie, Mirza, Xu, Warde-Farley, Ozair, Courville, and Bengio]{goodfellow2014generative}
Goodfellow, I., Pouget-Abadie, J., Mirza, M., Xu, B., Warde-Farley, D., Ozair, S., Courville, A., and Bengio, Y.
\newblock Generative adversarial nets.
\newblock \emph{NeurIPS}, 2014.

\bibitem[He et~al.(2016)He, Zhang, Ren, and Sun]{he2016deep}
He, K., Zhang, X., Ren, S., and Sun, J.
\newblock Deep residual learning for image recognition.
\newblock In \emph{CVPR}, 2016.

\bibitem[Heusel et~al.(2017)Heusel, Ramsauer, Unterthiner, Nessler, and Hochreiter]{heusel2017gans}
Heusel, M., Ramsauer, H., Unterthiner, T., Nessler, B., and Hochreiter, S.
\newblock Gans trained by a two time-scale update rule converge to a local nash equilibrium.
\newblock \emph{NeurIPS}, 2017.

\bibitem[Ho et~al.(2020)Ho, Jain, and Abbeel]{ho2020denoising}
Ho, J., Jain, A., and Abbeel, P.
\newblock Denoising diffusion probabilistic models.
\newblock \emph{NeurIPS}, 2020.

\bibitem[Ioffe \& Szegedy(2015)Ioffe and Szegedy]{ioffe2015batch}
Ioffe, S. and Szegedy, C.
\newblock Batch normalization: Accelerating deep network training by reducing internal covariate shift.
\newblock In \emph{ICML}, 2015.

\bibitem[Kingma \& Welling(2013)Kingma and Welling]{kingma2013auto}
Kingma, D.~P. and Welling, M.
\newblock Auto-encoding variational bayes.
\newblock \emph{arXiv preprint arXiv:1312.6114}, 2013.

\bibitem[Krizhevsky et~al.(2009)Krizhevsky, Hinton, et~al.]{krizhevsky2009learning}
Krizhevsky, A., Hinton, G., et~al.
\newblock Learning multiple layers of features from tiny images.
\newblock 2009.

\bibitem[Liu et~al.(2015)Liu, Luo, Wang, and Tang]{liu2015deep}
Liu, Z., Luo, P., Wang, X., and Tang, X.
\newblock Deep learning face attributes in the wild.
\newblock In \emph{ICCV}, 2015.

\bibitem[Loshchilov \& Hutter(2017)Loshchilov and Hutter]{loshchilov2017decoupled}
Loshchilov, I. and Hutter, F.
\newblock Decoupled weight decay regularization.
\newblock \emph{arXiv preprint arXiv:1711.05101}, 2017.

\bibitem[Paszke et~al.(2019)Paszke, Gross, Massa, Lerer, Bradbury, Chanan, Killeen, Lin, Gimelshein, Antiga, et~al.]{paszke2019pytorch}
Paszke, A., Gross, S., Massa, F., Lerer, A., Bradbury, J., Chanan, G., Killeen, T., Lin, Z., Gimelshein, N., Antiga, L., et~al.
\newblock Pytorch: An imperative style, high-performance deep learning library.
\newblock \emph{NeurIPS}, 2019.

\bibitem[Radford et~al.(2015)Radford, Metz, and Chintala]{radford2015unsupervised}
Radford, A., Metz, L., and Chintala, S.
\newblock Unsupervised representation learning with deep convolutional generative adversarial networks.
\newblock \emph{arXiv preprint arXiv:1511.06434}, 2015.

\bibitem[Salimans et~al.(2016)Salimans, Goodfellow, Zaremba, Cheung, Radford, and Chen]{salimans2016improved}
Salimans, T., Goodfellow, I., Zaremba, W., Cheung, V., Radford, A., and Chen, X.
\newblock Improved techniques for training gans.
\newblock \emph{NeurIPS}, 2016.

\bibitem[Saseendran et~al.(2021)Saseendran, Skubch, Falkner, and Keuper]{saseendran2021shape}
Saseendran, A., Skubch, K., Falkner, S., and Keuper, M.
\newblock Shape your space: A gaussian mixture regularization approach to deterministic autoencoders.
\newblock \emph{NeurIPS}, 2021.

\bibitem[Van Den~Oord et~al.(2017)Van Den~Oord, Vinyals, et~al.]{van2017neural}
Van Den~Oord, A., Vinyals, O., et~al.
\newblock Neural discrete representation learning.
\newblock \emph{NeurIPS}, 2017.

\bibitem[Vincent et~al.(2008)Vincent, Larochelle, Bengio, and Manzagol]{vincent2008extracting}
Vincent, P., Larochelle, H., Bengio, Y., and Manzagol, P.-A.
\newblock Extracting and composing robust features with denoising autoencoders.
\newblock In \emph{ICML}, 2008.

\bibitem[Zhang et~al.(2018)Zhang, Isola, Efros, Shechtman, and Wang]{zhang2018unreasonable}
Zhang, R., Isola, P., Efros, A.~A., Shechtman, E., and Wang, O.
\newblock The unreasonable effectiveness of deep features as a perceptual metric.
\newblock In \emph{CVPR}, 2018.

\end{thebibliography}
\bibliographystyle{icml2024}

\end{document}